\documentclass[conference]{IEEEtran}
\IEEEoverridecommandlockouts
\usepackage{cite}
\usepackage{amsmath,amssymb,amsfonts}
\usepackage{algorithmic}
\usepackage{graphicx}
\usepackage{placeins}
\usepackage{textcomp}
\usepackage{xcolor}

\def\BibTeX{{\rm B\kern-.05em{\sc i\kern-.025em b}\kern-.08em
T\kern-.1667em\lower.7ex\hbox{E}\kern-.125emX}}

\begin{document}

\title{Adaptive Multi-Expert Reasoning via Difficulty-Aware Routing and Uncertainty-Guided Aggregation}
\author{
\IEEEauthorblockN{Mohamed Ehab}
\IEEEauthorblockA{\textit{Faculty of Computer Science} \\
\textit{October University for Modern Science \& Arts}\\
Giza, Egypt \\
me338484@gmail.com}
\and
\IEEEauthorblockN{Ali Hamdi}
\IEEEauthorblockA{\textit{Faculty of Computer Science} \\
\textit{October University for Modern Science \& Arts}\\
Giza, Egypt \\
ahamdi@msa.edu.eg}
}
\maketitle

\begin{abstract}
Large language models (LLMs) demonstrate strong performance in math reasoning benchmarks, but their performance varies inconsistently across problems with varying levels of difficulty. This paper describes Adaptive Multi-Expert Reasoning (AMR), a framework that focuses on problem complexity by reasoning with dynamically adapted strategies. An agile routing system that focuses on problem text predicts problems' difficulty and uncertainty and guides a reconfigurable sampling mechanism to manage the breadth of generation. Three specialized experts create candidate responses, which are modified during multiple correction and finalization phases. A neural verifier assesses the correctness of responses, while a clustering-based aggregation technique identifies the final candidate answer based on a combination of consensus and answer quality. When evaluated on the GSM8K dataset, AMR achieved 75.28\% accuracy while only using the original training data. This result outperformed the majority of comparable 7B models that were trained on synthetic data. This showcases that models using difficulty-based routing and uncertainty-driven aggregation are efficient and effective in improving math reasoning models' robustness.
\end{abstract}


\section{Introduction}
There has been a lot of recent success with large language models (LLMs) and their ability to reason with mathematics, particularly with newly developed benchmarks like GSM8K \cite{cobbe2021trainingverifierssolvemath}. A lot of positive work has also been done with scaling, instruction tuning, and reasoning via chain-of-thought \cite{10.5555/3600270.3602070} for better performance with multi-step reasoning tasks. LLMs, however, still show a lot of variability with the problems that have different levels of reasoning complexity. They end up struggling to generalize, whether it be a simple arithmetic calculation, or a more difficult multi-step reasoning problem. Existing approaches like uniform prompting, a way to solve a problem without reasoning, and static ensemble averaging outputs, without problem difficulty and expert specialization, lack flexibility with 2 observations that must be considered: (1) more than one reasoning style is needed, and (2) problems can have varying levels of underlying complexity which can be difficult to solve.In this paper, we present AMR, a reasoning framework with multi-experts that is also aware of the problem’s difficulty, to solve these reasoning problems. The framework has 4 main components:  (i) a difficulty-aware router, (ii) multiple stylistically specialized experts, (iii) a neural verifier, and (iv) a clustering-based aggregation mechanism.

\begin{figure}[htbp]
\centering
\includegraphics[width=0.8\columnwidth]{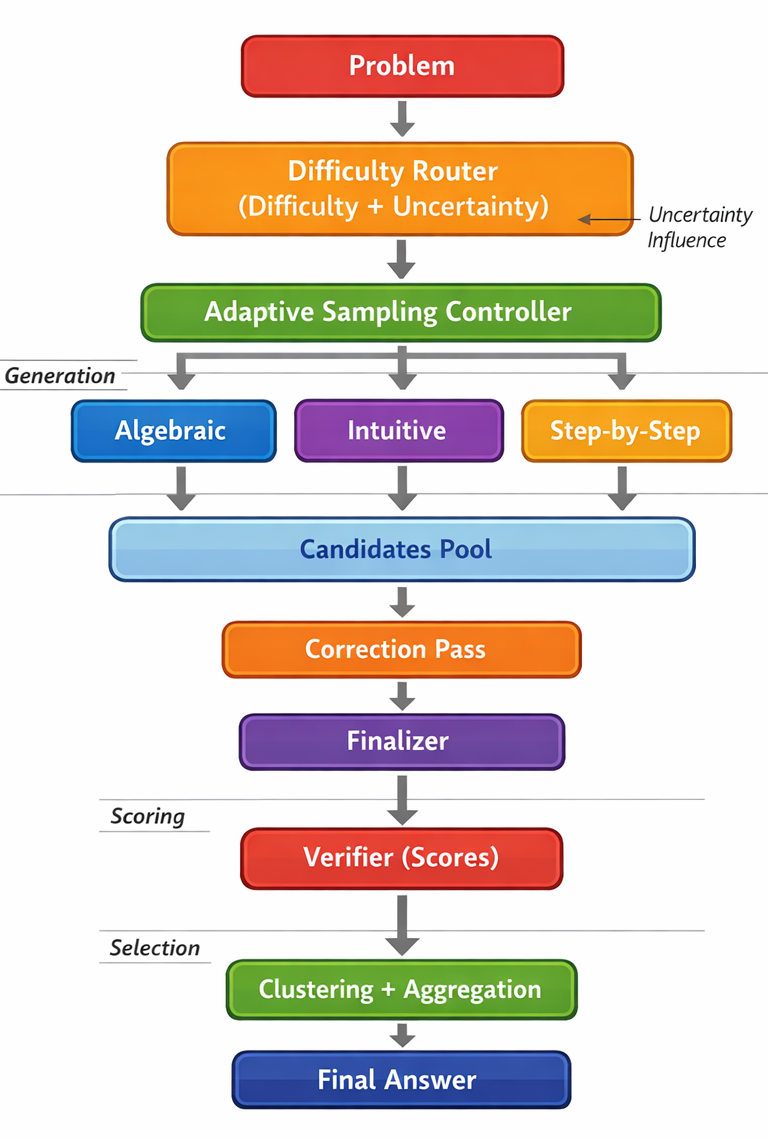}
\caption{AMR processes each problem based on predicted difficulty and uncertainty, routes them to various potential solution generators in different expert categories, and guides answer selection via correction and finalization to answer clusters as determined by a verifier.}
\label{fig:overview}
\end{figure}
\FloatBarrier
Our method has notable similarities to mixture-of-experts (MoE) architectures \cite{DBLP:conf/iclr/ShazeerMMDLHD17,JMLR:v23:21-0998}, though they are fundamentally different because they perform routing at inference time across reasoning styles, as opposed to through learned parameter sparsity in a single model. The router predicts both difficulty and uncertainty, which gives it the ability to control generation diversity adaptively. While in training and evaluation, the problem difficulty is estimated via a heuristic based on the number of reasoning steps in the reference solution (this serves as a stand-in for gold difficulty), during inference, the router model solely predicts difficulty from the problem text. This is to ensure that routing choices are made based on test-time evidence only. Three reasoning-expert models based on LoRA generate different reasoning paradigms. They provide additional candidates through correction and finalization passes, thus enhancing robustness. One verifier evaluates the candidates for correctness and a clustering-based aggregation method evaluates and selects the final output based on agreement and quality. Unlike most existing approaches, AMR does not depend on extensive synthetic data or enormous model size increases. Instead, it enhances reasoning through adaptive inference decision making. We identify four key contributions. First, we implement a difficulty-sensitive routing mechanism featuring uncertainty estimation that directly manages the number and variety of reasoning approaches generated.We also present a multi-expert reasoning framework that incorporates stylistically specialized LoRA-adapted experts, as well as a correction step that improves erroneous answers and a finalization step that delivers clear and high-quality outputs. Third, we build a clustering-based aggregation approach that combines verifier confidence, answer quality, and expert consensus on how to reliably choose a final answer. To conclude, we show that our method outperforms competitors on data efficiency by achieving strong performance (75.28\% accuracy on GSM8K) using only the original training data. This is in contrast to other methods that depend on extensive, synthetic data augmentation.

\section{Related Work}
\subsection{Robustness Benchmarks and the GSM-PLUS Study}
GSM8K \cite{cobbe2021trainingverifierssolvemath} is a widely accepted metric to measure mathematical reasoning abilities of models. It has, however, been shown that models that achieve high accuracy on these tasks, still fail when the data is altered. Li et al. \cite{li2024gsmpluscomprehensivebenchmarkevaluating} present GSM-PLUS, a more complete benchmark with problem alterations from multiple directions (linguistic variations, numerical variations, alterations in the solution space). Here we present their most notable findings:

\begin{itemize}
\item \textbf{Closed-source models:} GPT-4 sees an 8.23\% absolute drop from 93.25\% on GSM8K to 85.58\% on GSM-PLUS; GPT-3.5  has a 16.88\% drop from 73.62\% to 61.19\%.
\item \textbf{Open-source base models:} Mistral-7B sees a drop from 39.58\% to 26.18\%; LLaMA-2-7B from 13.42\% to 8.12\%; LLaMA-2-70B from 56.71\% to 40.04\%.
\item \textbf{Math-specialized models:} \textit{Even with fine-tuning on synthetic data,} MetaMath-7B drops from 66.72\% to 44.35\%; ToRA (7B) from 67.48\% to 43.60\%; MAmmoTH (7B) from 52.84\% to 32.14\%.
\end{itemize}

Robustness yet is a crucial concern, and this is what triggered our emphasis on uncertainty modeling, and difficulty-aware routing.

\subsection{Data Scaling and Verifier-Based Selection}
An alternative line of research examines the limits of small models trained with plentiful synthetic data. Approaches like ReAct \cite{unknown2} combine reasoning and external tool usage. Augmented problem-solving abilities have been shown to be the result of operational language models. Program-aided methods like PAL \cite{10.5555/3618408.3618843} use symbolic execution to enhance their numerical precision, and this is seen as a neural reasoning alternative. Liu et al. \cite{liu2023tinygsmachieving80gsm8k} introduce TinyGSM, a dataset of 12.3 million grade school math problems with corresponding Python solutions derived from a superior teacher model. After refinement on TinyGSM, a pair of Phi-1.5 models (1.3B each) – a generator and a verifier – achieves 81.5\% accuracy on GSM8K, surpassing significantly larger baselines. Some of the most important results of their work includes:

\begin{itemize}

    \item \textbf{Baselines:} GPT-3.5 (77.4\%), GPT-4 (97.0\%); LLaMA-2-7B (14.6\%), LLaMA-2-70B (56.8\%), Mistral-7B (52.2\%).

    \item \textbf{Math-tuned models:} MetaMath-7B (66.5\%), MetaMath-70B (82.3\%), WizardMath-70B (81.6\%), ToRA-Code-70B (84.3\%).

    \item \textbf{TinyGSM progression:} Phi-1.5 (44.6\%), Phi-GSM (68.2\%), Phi-GSM+V (81.5\%).

\end{itemize}

These findings demonstrate the advantages of integrating large-scale synthetic data with verifier-guided selection, even though the data needs are high. According to \cite{unknown}, self-consistency decoding improves results through the use of different reasoning pathways and the selection of consistent results. In contrast, AMR is data efficient, achieving 75.28\% accuracy with only the original GSM8K dataset. In addition, AMR is built on a 7B-scale model, whereas closed-source models like GPT-3.5 and GPT-4 are assumed to operate on a significantly higher level, often in the tens to hundreds of billions of parameters. This model capacity variation should be taken into account in performance comparisons.

\subsection{Enhanced Problem Understanding (DUP)}
Zhong et al. \cite{Zhong2024AchievingO} state that chain-of-thought prompting has issues with semantics. As a result, they created DUP, a zero-shot method that aims to isolate and extract the main question, the question’s problem-solving info, and reasoning. Their findings are impressive:

\begin{itemize}

    \item DUP with GPT-4 is 97.1\% on accuracy on GSM8K (94.6\% for Zero-shot CoT).

    \item SVAMP accuracy increases from 90.4\% to 94.2\%.

    \item An error analysis of DUP shows a reduction of semantic errors from 54 to 33, calculation errors from 18 to 9, and step-missing errors from 10 to 2.

\end{itemize}

DUP shows that increased understanding of the problem is the main driving factor behind the increase in performance, and this can be achieved without any further fine-tuning. Our approach focuses on inference-time architecture (routing, diversity, and aggregation) as opposed to prompt engineering. With this understanding in the modules and our framework, we can expect to achieve greater results.

\subsection{Positioning of Our Work}
Prior research has cultivated mathematical reasoning along three dimensions:
\begin{enumerate}
    \item \textbf{Robustness benchmarking (GSM-PLUS)}: shows models have significant drops in accuracy due to distribution shifts.

    \item \textbf{Data scaling (TinyGSM)}: shows how small models can reach high accuracy when trained on a large synthetic dataset.

    \item \textbf{Problem understanding (DUP)}: shows how the right prompt can significantly boost zero-shot performance.
\end{enumerate}
  
With this work, we add a new dimension: \textbf{inference-time architecture}. We demonstrate that mechanisms like difficulty aware routing, multi-expert diversity, and uncertainty-aggregation can perform well using only the original training data without any synthetic dataset augmentation or costly model scaling. This underscores the value of sophisticated inference mechanisms, separate and in addition to, data scaling and prompt techniques.

\section{Methodology}
\subsection{Difficulty-Aware Router}
When faced with a specific problem $x$, the router generates a probability distribution concerning two classes of difficulty. The role of uncertainty in the decision-making process is significant \cite{10.5555/3295222.3295309}, which motivates our hybrid entropy–margin formulation. A hybrid uncertainty measure is given by:

\begin{equation}
U(x) = \frac{1}{2} H(p(x)) + \frac{1}{2} \left(1 - 2\left| p_{\text{hard}}(x) - 0.5 \right| \right),
\end{equation}

\begin{figure}[htbp]

\centering

\includegraphics[width=\columnwidth]{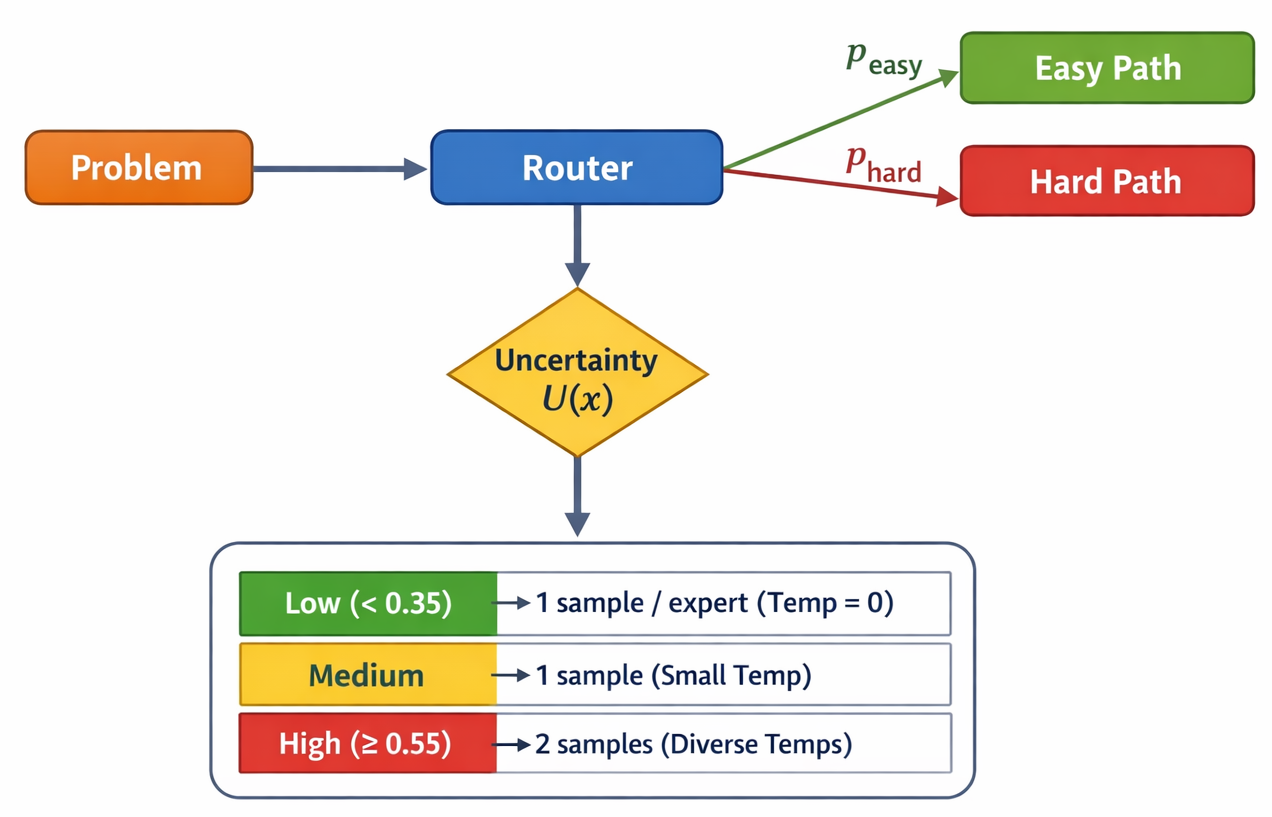}

\caption{Generated strategy based on uncertainty and difficulty. The router estimates problem difficulty and computes a hybrid uncertainty score, which influences how many diverse candidates will be generated.}

\label{fig:overview}

\end{figure}

\FloatBarrier

Where $H(p(x))$ is the Shannon entropy. Based on $U(x)$, we can describe 3 regimes:

\begin{itemize}

\item Low uncertainty ($U<0.35$): → deterministic generation.

\item Medium uncertainty ($0.35\le U<0.55$): → one candidate per expert with a low temperature.

\item High uncertainty ($U\ge 0.55$): → two candidates per expert, with temperatures 0.0 and 0.15.

\end{itemize}

\subsection{Multi-Expert Reasoning}
Our system employs three LoRA-adapted specialists, each trained with different prompts:

\begin{itemize}

\item Algebraic: equation-based reasoning.

\item Intuitive: mental math and natural language.

\item Step-by-step: detailed structured line-by-line derivations.

\end{itemize}

Additionally, we propose:

\begin{itemize}

\item \textbf{Correction pass}: The step-by-step expert tries to correct the first mistake in the best candidate(s). Our correction pass is similar to the correction stage in Self-Refine \cite{10.5555/3666122.3668141} where models refine their output based on feedback.

\item \textbf{Finalizer pass}: A short and high-quality solution produced by the step-by-step expert.

\end{itemize}

We filter all candidates to keep only those which answer the question and provide a clear answer indicator.

\begin{figure}[htbp]

\centering

\includegraphics[width=0.9\columnwidth]{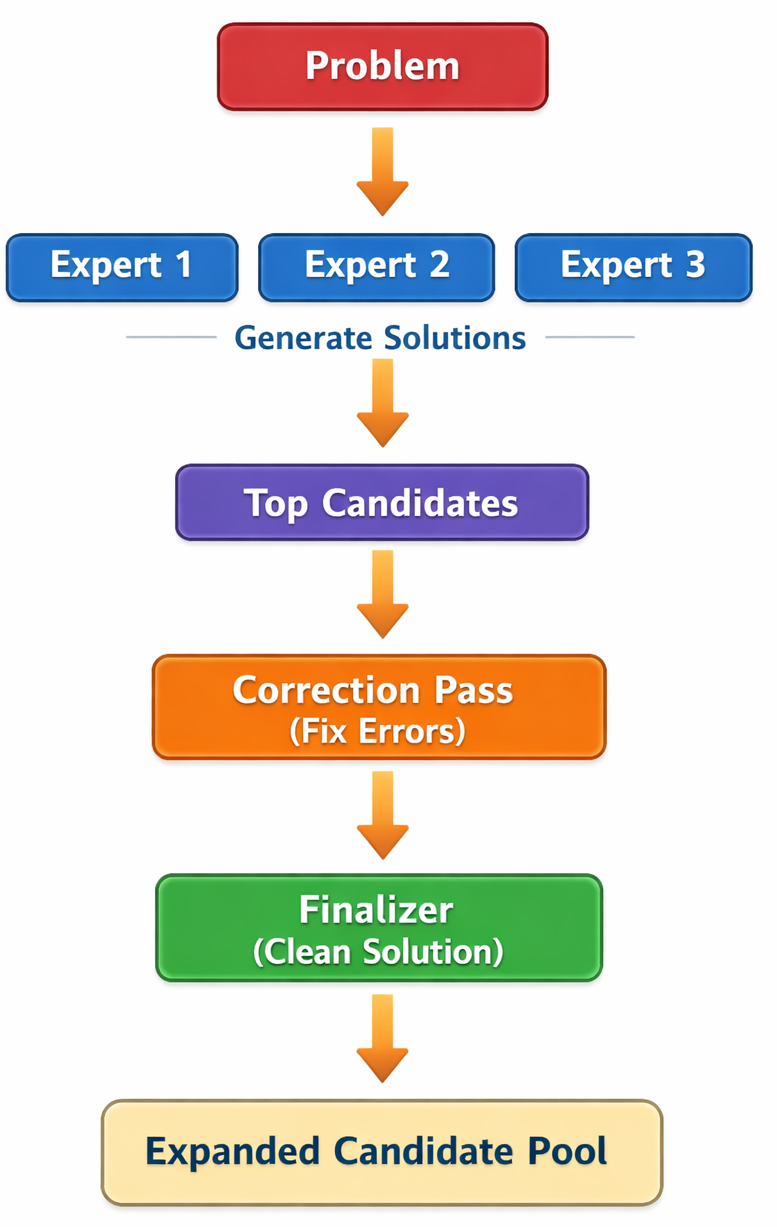}

\caption{Multi-expert reasoning pipeline. Specialized experts generate diverse solutions, which are refined through a correction stage and a finalization stage to improve accuracy and clarity.}

\label{fig:overview}

\end{figure}
\FloatBarrier

\subsection{Neural Verifier}
The verifier operates as a binary classifier (DeBERTa-v3) which has been trained on concatenated pairs of problems and their respective solutions. This method is documented in the earlier studies which focus on training verifiers in the field of mathematical reasoning \cite{cobbe2021trainingverifierssolvemath}. Specifically, it has been demonstrated that solution accuracy is improved significantly when employing verifier-guided selection. Numerical answer matching allows for the categorization of each candidate as being either correct or incorrect. The verifier assigns a score between 0 and 1 inclusive which represents the likelihood of the answer's correctness.

\subsection{Clustering-Based Aggregation}
All applicants are assigned a combined score:

\begin{equation}
\text{score} = 0.50\,s_{\text{verifier}} + 0.18\,c_{\text{completion}} + 0.16\,q_{\text{quality}} + 0.16\,b_{\text{source}},
\end{equation}

The weights were empirically adjusted on a small validation subset in order to balance the verifier confidence against the heuristic signals. The definitions are:

\begin{itemize}
    \item $c_{\text{completion}}$ is a heuristic bonus for nicely structured answers. $c_{\text{completion}} = 0.30$ if the answer includes ``\#\#\#\#'', plus $0.10$ if it includes ``final answer'' (max $0.40$), and minus $0.15$ if it contains code fences.
    \item $q_{\text{quality}}$ represents the coherence of the answer. It is the normalized score of $(\text{len(answer)} / 200)$ clamped to $[0,1]$ and penalized by $ -0.12$ if a code keyword was present.
    \item $b_{\text{source}}$ gives specific generation passes bonuses: $0.22$ for finalizer candidates, $0.14$ for corrector candidates and $0$ otherwise.
\end{itemize}

Candidates are grouped by the numerical answer extracted. Our clustering method is inspired by ensemble decision theory \cite{6410720} that integrates diversity and consensus. For each cluster $c$, the score of the cluster is

\begin{equation}
\begin{aligned}
    \text{cluster\_score} = &\; 0.42 \cdot \max_{m \in c} \text{score}(m) + 0.16 \cdot \bar{\text{score}}_c \\
    & + 0.10 \cdot \min\!\left(\frac{\text{expert\_support}_c}{3}, 1\right) \\
    & + 0.10 \cdot \min\!\left(\frac{|c|}{4}, 1\right)
\end{aligned}
\end{equation}

The final output is the best candidate in the best cluster.

\begin{figure}[htbp]
\centering
\includegraphics[width=\columnwidth]{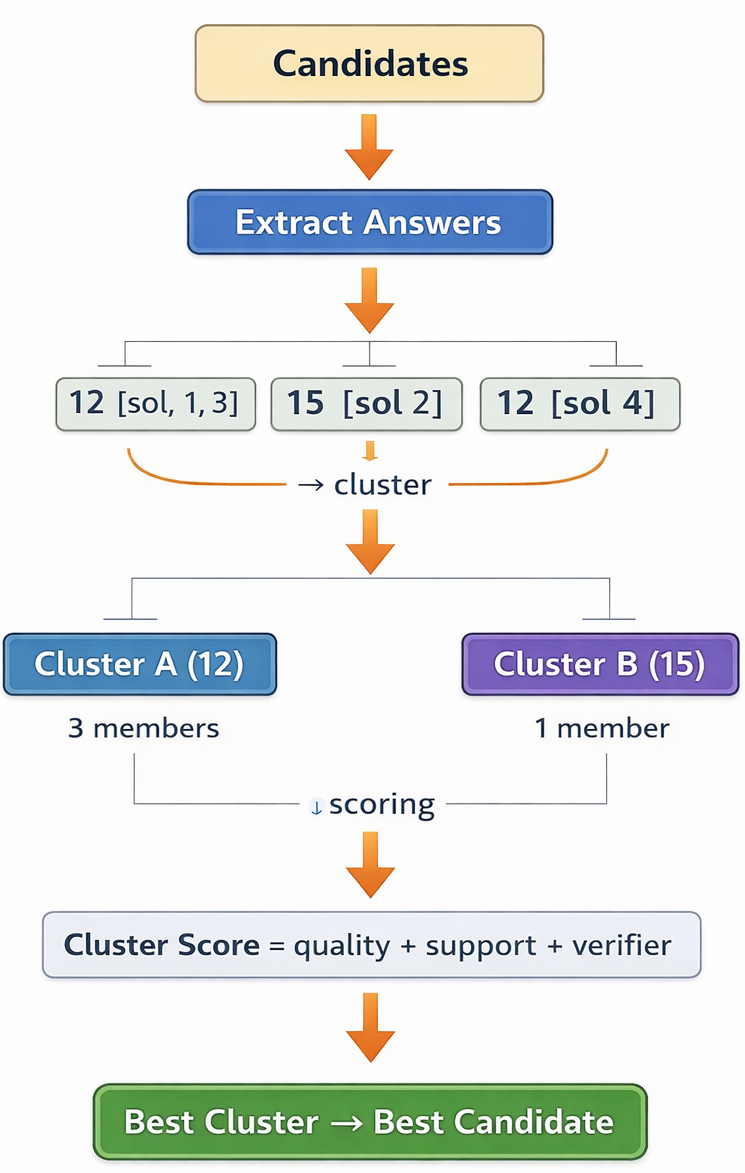}
\caption{Aggregation by clustering. Candidate answers are grouped by the extracted answers and the clusters are scored based on verifier confidence, quality, and consensus. The best answer is selected from the best cluster.}
\label{fig:overview}
\end{figure}
\FloatBarrier

\section{Results}
We assess AMR on the GSM8K test split (1,319 examples). The results are summarized in Table~\ref{tab:accuracy}.

\begin{table}[htbp]
\centering
\caption{AMR accuracy on GSM8K test split.}
\label{tab:accuracy}
\begin{tabular}{lccc}
\hline
\textbf{Difficulty} & \textbf{Correct} & \textbf{Total} & \textbf{Accuracy} \\
\hline
Easy (predicted) & 657 & 795 & 82.6\% \\
Hard (predicted) & 336 & 524 & 64.1\% \\
\hline
Easy (gold) & 636 & 742 & 85.7\% \\
Hard (gold) & 357 & 577 & 61.9\% \\
\hline
Overall & 993 & 1319 & \textbf{75.28\%} \\
\hline
\end{tabular}
\end{table}

\subsection{Comparison with Prior Work}
We evaluate AMR in comparison to relevant baseline and state-of-the-art models. Models labeled with $\dagger$ have used further synthetic data beyond the original GSM8K training set.

\begin{table}[htbp]
\centering
\footnotesize
\setlength{\tabcolsep}{3pt}
\caption{Models without mathematical fine-tuned specific data on GSM8K. Different evaluation methods are used (e.g. pass@1, self-consistency, or few-shot prompting) and thus not directly comparable.
}
\label{tab:foundation}
\begin{tabular}{lccc}
\hline
\textbf{Model} & \textbf{Base Model} & \textbf{Size} & \textbf{GSM8K (\%)} \\
\hline
LLaMA-2 \cite{touvron2023llama2openfoundation} & – & 7B & 14.6 \\
LLaMA-2 & – & 13B & 28.7 \\
LLaMA-2 & – & 34B & 42.2 \\
LLaMA-2 & – & 70B & 56.8 \\
Mistral \cite{Jiang2023Mistral7} & – & 7B & 52.2 \\
Mistral & – & 8$\times$7B & 58.4 \\
CodeLlama \cite{yue2024mammoth} & LLaMA-2 & 7B & 25.3 \\
CodeLlama & LLaMA-2 & 13B & 35.9 \\
CodeLlama & LLaMA-2 & 34B & 45.6 \\
LLaMA-2 (8-shot) \cite{li2024gsmpluscomprehensivebenchmarkevaluating} & – & 7B & 13.4 \\
LLaMA-2 (8-shot) & – & 13B & 25.4 \\
LLaMA-2 (8-shot) & – & 70B & 56.7 \\
Mistral (8-shot) \cite{li2024gsmpluscomprehensivebenchmarkevaluating} & – & 7B & 39.6 \\
\hline
\end{tabular}
\end{table}

\begin{table}[htbp]
\centering
\footnotesize
\setlength{\tabcolsep}{1pt}
\caption{Math‑tuned models on GSM8K. $\dagger$ indicates use of additional synthetic data.}
\label{tab:math_tuned}
\resizebox{\columnwidth}{!}{%
\begin{tabular}{lcccc}
\hline
\textbf{Model} & \textbf{Base Model} & \textbf{Size} & \textbf{Synthetic Data} & \textbf{GSM8K(\%)} \\
\hline
MetaMath $\dagger$ \cite{Yu2023MetaMathBY} & LLaMA-2 & 7B & MetaMathQA (395K) & 66.7 \\
MetaMath $\dagger$ & LLaMA-2 & 13B & MetaMathQA (395K) & 70.8 \\
MetaMath $\dagger$ & LLaMA-2 & 70B & MetaMathQA (395K) & 82.1 \\
MetaMath-Mistral $\dagger$ & Mistral & 7B & MetaMathQA (395K) & 77.8 \\
WizardMath $\dagger$ \cite{luo2025wizardmathempoweringmathematicalreasoning} & LLaMA-2 & 7B & Evol‑Instruct & 54.9 \\
WizardMath $\dagger$ & LLaMA-2 & 13B & Evol‑Instruct & 63.9 \\
WizardMath $\dagger$ & LLaMA-2 & 70B & Evol‑Instruct & 81.6 \\
ToRA-Code $\dagger$ \cite{Gou2023ToRAAT} & LLaMA-2 & 7B & TORA‑CORPUS (16K) & 72.6 \\
ToRA-Code $\dagger$ & LLaMA-2 & 13B & TORA‑CORPUS (16K) & 75.8 \\
ToRA-Code $\dagger$ & LLaMA-2 & 34B & TORA‑CORPUS (16K) & 80.7 \\
ToRA-Code $\dagger$ & LLaMA-2 & 70B & TORA‑CORPUS (16K) & 84.3 \\
OVM $\dagger$ & LLaMA-2 & 7B+7B & – & 73.7 \\
OVM $\dagger$ & Mistral & 7B+7B & – & 84.7 \\
MAMmoTH $\dagger$ \cite{yue2024mammoth} & LLaMA-2 & 7B & MathInstruct (260K) & 52.8 \\
MAMmoTH $\dagger$ & LLaMA-2 & 13B & MathInstruct (260K) & 62.4 \\
MAMmoTH $\dagger$ & LLaMA-2 & 70B & MathInstruct (260K) & 75.9 \\
MAMmoTH-Coder $\dagger$ & CodeLLaMA & 7B & MathInstruct (260K) & 59.9 \\
MAMmoTH-Coder $\dagger$ & CodeLLaMA & 13B & MathInstruct (260K) & 64.9 \\
SEGO $\dagger$ & CodeLLaMA & 7B & GSM8K+MATH+AQuA & 68.7 \\
SEGO $\dagger$ & CodeLLaMA & 13B & GSM8K+MATH+AQuA & 72.5 \\
Abel $\dagger$ & LLaMA-2 & 7B & Unreleased & 59.5 \\
Abel $\dagger$ & LLaMA-2 & 13B & Unreleased & 66.7 \\
Abel $\dagger$ & LLaMA-2 & 70B & Unreleased & 83.9 \\
Phi-GSM $\dagger$ \cite{liu2023tinygsmachieving80gsm8k} & Phi-1.5 & 1.3B & TinyGSM (1.3M) & 68.2 \\
Phi-GSM+V $\dagger$ & Phi-1.5 & 1.3B+1.3B & TinyGSM (1.3M) & 81.5 \\
Phi-GSM (125M) $\dagger$ & Phi-1.5-tiny & 125M & TinyGSM (1.3M) & 63.1 \\
Phi-GSM+V (125M) $\dagger$ & Phi-1.5-tiny & 125M+125M & TinyGSM (1.3M) & 68.9 \\
Phi-GSM (350M) $\dagger$ & Phi-1.5-small & 350M & TinyGSM (1.3M) & 65.9 \\
Phi-GSM+V (350M) $\dagger$ & Phi-1.5-small & 350M+350M & TinyGSM (1.3M) & 71.3 \\
Phi-GSM $\dagger$ & Phi-2 & 2.7B & TinyGSM (1.3M) & 74.3 \\
\hline
\multicolumn{5}{c}{\textit{Proposed Method (AMR)}} \\
\hline
\textbf{(AMR)} & Qwen2.5-Math & 7B & \textbf{No} & \textbf{75.28} \\
\hline
\end{tabular}
}
\end{table}
\FloatBarrier
AMR is the first model to achieve 75.28\% accuracy on the original GSM8K training data, surpassing the majority of 7B-scale models with synthetic data (MetaMath, WizardMath, and ToRA) and coming closer to the 13B model systems. Interestingly, Phi-GSM+V utilizes more data (over 1.3M synthetic examples plus an extra verifier model) to achieve higher accuracy (81.5\%). These results emphasize the effectiveness and efficiency of the data, difficulty-aware routing, and uncertainty-guided aggregation.

\subsection{Implications for Robustness and Orthogonality}
Results from GSM-PLUS suggest that model accuracy drops significantly due to distribution shifts, regardless of how extensively the model has been trained on a large synthetic dataset. Because our approach does not depend on synthetic data, it may display a different type of robustness due to the diversity-driven inference. While we have not evaluated our model on GSM-PLUS, we believe that the approaches employed in this model of multiple experts, correction passes, and clustering are best suited for variety in the different problem contexts. Additionally, our approach differs from DUP in that we focus on diversity of inference, while DUP relies on prompt engineering to improve the understanding of the problem. When used together, DUP-style preprocessing and our inference architecture have the potential to improve performance significantly. We posit that DUP reduces semantic and step-missing errors while our framework stabilizes the choice of reasoning path.

\section{Discussion}
The router's difficulty classification achieves 73.4\% alignment with the gold standard on the test set. The uncertainty band impacts generation diversity, and this plays a role in the 64.1\% accuracy achieved on hard problems—remarkably better than the result obtained from a single deterministic run.The verifier delivers trustworthy estimates of correctness, and the clustering mechanism provides stability in selection—especially in situations where multiple experts give different outputs. The high-quality candidates added by the correction and finalizer passes are captured in the scoring formula by their bonuses. Limitations include the router’s accuracy (which could improve with additional training), a static set of three experts, and evaluation restricted to GSM8K. Broader applicability could be shown by extending to additional benchmarks such as MATH and SVAMP, and future work should evaluate GSM-PLUS for robustness as well.

\section{Conclusion}
We introduced AMR, a multi-expert reasoning framework that is aware of difficulty and incorporates adaptive routing, specialized experts, a neural verifier, and clustering-based aggregation. It attained 75.28\% accuracy on GSM8K, using solely the original training data, exceeding the performance of many models that are based on large-scale synthetic augmentation. Our findings suggest that smart inference-time strategies can be just as important as data augmentation. We will pursue dynamic expert selection, enhanced uncertainty modeling, and broader reasoning domains, as well as robustness evaluation on the perturbed benchmark GSM-PLUS.

\bibliographystyle{plain}
\bibliography{mybibfile}

\begin{thebibliography}{10}

\bibitem{cobbe2021trainingverifierssolvemath}
Karl Cobbe, Vineet Kosaraju, Mohammad Bavarian, Mark Chen, Heewoo Jun, Lukasz Kaiser, Matthias Plappert, Jerry Tworek, Jacob Hilton, Reiichiro Nakano, Christopher Hesse, and John Schulman.
\newblock Training verifiers to solve math word problems, 2021.

\bibitem{JMLR:v23:21-0998}
William Fedus, Barret Zoph, and Noam Shazeer.
\newblock Switch transformers: Scaling to trillion parameter models with simple and efficient sparsity.
\newblock {\em Journal of Machine Learning Research}, 23(120):1--39, 2022.

\bibitem{10.5555/3618408.3618843}
Luyu Gao, Aman Madaan, Shuyan Zhou, Uri Alon, Pengfei Liu, Yiming Yang, Jamie Callan, and Graham Neubig.
\newblock Pal: program-aided language models.
\newblock In {\em Proceedings of the 40th International Conference on Machine Learning}, ICML'23. JMLR.org, 2023.

\bibitem{Gou2023ToRAAT}
Zhibin Gou, Zhihong Shao, Yeyun Gong, Yelong Shen, Yujiu Yang, Minlie Huang, Nan Duan, and Weizhu Chen.
\newblock Tora: A tool-integrated reasoning agent for mathematical problem solving.
\newblock {\em ArXiv}, abs/2309.17452, 2023.

\bibitem{Jiang2023Mistral7}
Albert~Qiaochu Jiang, Alexandre Sablayrolles, Arthur Mensch, Chris Bamford, Devendra~Singh Chaplot, Diego de~Las~Casas, Florian Bressand, Gianna Lengyel, Guillaume Lample, Lucile Saulnier, L{\'e}lio~Renard Lavaud, Marie-Anne Lachaux, Pierre Stock, Teven~Le Scao, Thibaut Lavril, Thomas Wang, Timoth{\'e}e Lacroix, and William~El Sayed.
\newblock Mistral 7b.
\newblock {\em ArXiv}, abs/2310.06825, 2023.

\bibitem{10.5555/3295222.3295309}
Alex Kendall and Yarin Gal.
\newblock What uncertainties do we need in bayesian deep learning for computer vision?
\newblock In {\em Proceedings of the 31st International Conference on Neural Information Processing Systems}, NIPS'17, page 5580–5590, Red Hook, NY, USA, 2017. Curran Associates Inc.

\bibitem{li2024gsmpluscomprehensivebenchmarkevaluating}
Qintong Li, Leyang Cui, Xueliang Zhao, Lingpeng Kong, and Wei Bi.
\newblock Gsm-plus: A comprehensive benchmark for evaluating the robustness of llms as mathematical problem solvers, 2024.

\bibitem{liu2023tinygsmachieving80gsm8k}
Bingbin Liu, Sebastien Bubeck, Ronen Eldan, Janardhan Kulkarni, Yuanzhi Li, Anh Nguyen, Rachel Ward, and Yi~Zhang.
\newblock Tinygsm: achieving >80

\bibitem{luo2025wizardmathempoweringmathematicalreasoning}
Haipeng Luo, Qingfeng Sun, Can Xu, Pu~Zhao, Jianguang Lou, Chongyang Tao, Xiubo Geng, Qingwei Lin, Shifeng Chen, Yansong Tang, and Dongmei Zhang.
\newblock Wizardmath: Empowering mathematical reasoning for large language models via reinforced evol-instruct, 2025.

\bibitem{10.5555/3666122.3668141}
Aman Madaan, Niket Tandon, Prakhar Gupta, Skyler Hallinan, Luyu Gao, Sarah Wiegreffe, Uri Alon, Nouha Dziri, Shrimai Prabhumoye, Yiming Yang, Shashank Gupta, Bodhisattwa~Prasad Majumder, Katherine Hermann, Sean Welleck, Amir Yazdanbakhsh, and Peter Clark.
\newblock Self-refine: iterative refinement with self-feedback.
\newblock In {\em Proceedings of the 37th International Conference on Neural Information Processing Systems}, NIPS '23, Red Hook, NY, USA, 2023. Curran Associates Inc.

\bibitem{6410720}
Friedhelm Schwenker.
\newblock Ensemble methods: Foundations and algorithms [book review].
\newblock {\em IEEE Computational Intelligence Magazine}, 8(1):77--79, 2013.

\bibitem{DBLP:conf/iclr/ShazeerMMDLHD17}
Noam Shazeer, Azalia Mirhoseini, Krzysztof Maziarz, Andy Davis, Quoc~V. Le, Geoffrey~E. Hinton, and Jeff Dean.
\newblock Outrageously large neural networks: The sparsely-gated mixture-of-experts layer.
\newblock In {\em 5th International Conference on Learning Representations, {ICLR} 2017, Toulon, France, April 24-26, 2017, Conference Track Proceedings}. OpenReview.net, 2017.

\bibitem{touvron2023llama2openfoundation}
Hugo Touvron, Louis Martin, Kevin Stone, Peter Albert, Amjad Almahairi, Yasmine Babaei, Nikolay Bashlykov, Soumya Batra, Prajjwal Bhargava, Shruti Bhosale, Dan Bikel, Lukas Blecher, Cristian~Canton Ferrer, Moya Chen, Guillem Cucurull, David Esiobu, Jude Fernandes, Jeremy Fu, Wenyin Fu, Brian Fuller, Cynthia Gao, Vedanuj Goswami, Naman Goyal, Anthony Hartshorn, Saghar Hosseini, Rui Hou, Hakan Inan, Marcin Kardas, Viktor Kerkez, Madian Khabsa, Isabel Kloumann, Artem Korenev, Punit~Singh Koura, Marie-Anne Lachaux, Thibaut Lavril, Jenya Lee, Diana Liskovich, Yinghai Lu, Yuning Mao, Xavier Martinet, Todor Mihaylov, Pushkar Mishra, Igor Molybog, Yixin Nie, Andrew Poulton, Jeremy Reizenstein, Rashi Rungta, Kalyan Saladi, Alan Schelten, Ruan Silva, Eric~Michael Smith, Ranjan Subramanian, Xiaoqing~Ellen Tan, Binh Tang, Ross Taylor, Adina Williams, Jian~Xiang Kuan, Puxin Xu, Zheng Yan, Iliyan Zarov, Yuchen Zhang, Angela Fan, Melanie Kambadur, Sharan Narang, Aurelien Rodriguez, Robert Stojnic, Sergey Edunov, and Thomas
  Scialom.
\newblock Llama 2: Open foundation and fine-tuned chat models, 2023.

\bibitem{unknown}
Xuezhi Wang, Jason Wei, Dale Schuurmans, Quoc Le, Ed~Chi, and Denny Zhou.
\newblock Self-consistency improves chain of thought reasoning in language models, 03 2022.

\bibitem{10.5555/3600270.3602070}
Jason Wei, Xuezhi Wang, Dale Schuurmans, Maarten Bosma, Brian Ichter, Fei Xia, Ed~H. Chi, Quoc~V. Le, and Denny Zhou.
\newblock Chain-of-thought prompting elicits reasoning in large language models.
\newblock In {\em Proceedings of the 36th International Conference on Neural Information Processing Systems}, NIPS '22, Red Hook, NY, USA, 2022. Curran Associates Inc.

\bibitem{unknown2}
Shunyu Yao, Jeffrey Zhao, Dian Yu, Nan Du, Izhak Shafran, Karthik Narasimhan, and Yuan Cao.
\newblock React: Synergizing reasoning and acting in language models, 10 2022.

\bibitem{Yu2023MetaMathBY}
Long~Long Yu, Weisen Jiang, Han Shi, Jincheng Yu, Zhengying Liu, Yu~Zhang, James~T. Kwok, Zheng Li, Adrian Weller, and Weiyang Liu.
\newblock Metamath: Bootstrap your own mathematical questions for large language models.
\newblock {\em ArXiv}, abs/2309.12284, 2023.

\bibitem{yue2024mammoth}
Xiang Yue, Xingwei Qu, Ge~Zhang, Yao Fu, Wenhao Huang, Huan Sun, Yu~Su, and Wenhu Chen.
\newblock Mammoth: Building math generalist models through hybrid instruction tuning.
\newblock In {\em International Conference on Learning Representations (ICLR)}, 2024.

\bibitem{Zhong2024AchievingO}
Qihuang Zhong, Kang Wang, Ziyang Xu, Juhua Liu, Liang Ding, Bo~Du, and Dacheng Tao.
\newblock Achieving >97\% on gsm8k: deeply understanding the problems makes llms better solvers for math word problems.
\newblock {\em Frontiers of Computer Science}, 20, 2024.

\end{thebibliography}

\end{document}